\documentclass[3p,times,procedia]{elsarticle}
\usepackage{ecrc}
\usepackage[bookmarks=false]{hyperref}
    \hypersetup{colorlinks,
      linkcolor=blue,
      citecolor=blue,
      urlcolor=blue}
\volume{00}

\firstpage{1}

\journalname{Procedia Computer Science}

\runauth{Mandischer, B\"{o}ckmann, Holl, Mikelsons}

\jid{procs}

\usepackage{amssymb}
\usepackage[figuresright]{rotating}
\usepackage{multirow}
\usepackage{subcaption}
\usepackage{array}

\usepackage[english]{babel}
\begin{document}
\begin{frontmatter}



\dochead{8th International Conference on Industry of the Future and Smart Manufacturing}%%%
%% e.g. \dochead{17th International Conference on Dynamical Processes in Excited States of Solids}

\title{ROS2SmolVLA: Enabling Small Vision-Language-Action Models for Integration into Industrial-Grade Lightweight Robots}

\author[a]{Nils Mandischer\fnref{equal}\corref{cor1}} 
\author[a]{Noah B\"{o}ckmann\fnref{equal}}
\author[a]{Ludwig Holl}
\author[a]{Lars Mikelsons}

\address[a]{University of Augsburg, Chair of Mechatronics, Am Technologiezentrum 8, 86159 Augsburg, Germany}

\begin{abstract}
Industrial demand changes the paradigms of production. Due to smaller batch sizes and more variations in products, companies face a growing challenge to adopt more adaptive production systems. In particular, robot-based automation is usually static and fails to respond to constantly changing processes. Vision-Language-Action (VLA) Models are a promising opportunity to mitigate this challenge by generating robot actions based on the observed system state. However, current research either focuses on large models that cannot be computed on premise, creating compliance and security challenges, or use lab-grade robot hardware that obscures exploitation in real industrial settings. In this work, we adapt Hugging~Face's SmolVLA for Universal Robots lightweight robots. Further, we release the open-source repository ROS2SmolVLA that implements an interface for ROS~2 to SmolVLA, and makes it applicable for industrial-grade hardware. By this, we allow a lenient adoption into lab and industrial environments. We validate the functionality of SmolVLA for a Universal Robots UR10e using a pick-and-place task and give implementation guidelines. Our findings support that SmolVLA is a well-suited option for small-sized tasks that need to be computed on premise.
\end{abstract}

\cortext[cor1]{Corresponding author. Tel.: +49 821 598 69287}
\fntext[equal]{Both authors contributed equally to this work.}
\end{frontmatter}

%\correspondingauthor[*]{Corresponding author. Tel.: +0-000-000-0000 ; fax: +0-000-000-0000.}
\email{nils.mandischer@uni-a.de}

% \linenumbers
\vspace*{-6pt}

%\enlargethispage{-7mm}
\section{Introduction}
\label{sec:introduction}
Modern industries face a paradigm change from rigid and sequential towards more flexible and adaptive production systems. These go hand in hand with smaller batch sizes, more variety in products, and move away from scheduled to on-demand production. While established automation systems work well in structured environments, flexible and adaptive production systems lead to more economic risks. In fact, a robot system that originally operated continually on a single work process must now be available for many processes and must adapt its execution to already minor changes in products.
In recent years, automation systems met this demand by establishing more adaptivity in planning and execution. Flexible Manufacturing Systems (FMS) provide more variety to changes in task sequence and batch sizes by adding more general purpose automation systems. In contrast, Reconfigurable Manufacturing Systems (RMS) use less hardware with a larger range of functionality and automation capabilities, leading to a lowered risk of over-automation. 
In recent years, Artificial Intelligence (AI) gained widespread use over the entire product life cycle. AI provides solutions for virtually all core concepts of RMS: modularity, integrability, convertibility, diagnosability, customization, and scalability~\cite{Koren2023,KombayaTouckia2022}. Thus RMS principles were extended into Smart Reconfigurable Manufacturing (SRM)~\cite{Bortolini2018}. SRM leverages the opportunities of AI capabilities in reconfigurable manufacturing, allowing for more adaptive planning and execution, and better abilities to react to unforeseen changes in the factory environment.

Reinforcement Learning (RL) has become a primary tool for SRM as it allows to handle high-dimensional data subject to uncertainty~\cite{Bahrpeyma2022}. RL gained traction with the introduction of digital twins into manufacturing, moving many planning and optimization operations from real-world into virtual environments. Recently, approaches like transformer-based RL using self-attention extended the abilities of production agents towards long-term temporal reasoning~\cite{Xu2024}.
With the quick rise of Large Language Models (LLMs), large-scale transformer models also got introduced into SRM~\cite{Li2026}. LLMs allow reasoning on big data without requiring trial-and-error testing like RL methods.
Recently, transformer-based Vision-Language-Action models (VLAs) emerged as a means of adaptively controlling robots~\cite{Kawaharazuka2025,Zitkovich2023}. VLAs analyze the status of the workspace through sensor-based observation and Vision-Language Models (VLMs). The tokenized results are then used by an Action Expert to generate robot control commands in real-time. VLAs are commonly trained on teleoperated demonstrations comprising of sensor data and robot control inputs.
LLMs and VLAs rely extensively on external compute and/or proprietary models, raising compliance and security concerns. Small VLAs, like Hugging~Face's SmolVLA~\cite{Shukor2025}, that feature significantly less parameters aim to relax these concerns by making LLMs available on premise or on the edge. However, their comparatively small context window prevents generalization about whole factory floors. Instead, these models must fulfill specific and bounded goals and actions.
Small VLAs are in the early stages of research and have not been made broadly available to real-world production systems. Instead, research relies on small-scale lab experiments with limited workspace dimensions and payloads.

In this work, we propose ROS2SmolVLA, a ROS~2 interface for SmolVLA that bridges the gap from lab experiments to real industrial-grade hardware. We make SmolVLA available for execution directly on Universal Robots lightweight robots with inference provided by edge devices and consumer-grade hardware comparable to edge PCs. We validate the applicability on an Universal Robots UR10e within a pick-and-place task, and indicate lessons learned from the deployment of SmolVLA in drastically larger workspaces.

The paper is structured as follows:
First, related work on VLAs is discussed (Section~\ref{sec:related_work}), including a detailed description of Hugging~Face's SmolVLA (Section~\ref{sec:smolvla}). In Section~\ref{sec:ROS~2smovla}, we present our open-source project ROS2SmolVLA that bridges ROS~2 and SmolVLA. Software and data are available via Github and Hugging~Face. Afterwards, we present a profound validation of ROS2SmolVLA with an Universal Robotis UR10e in a pick-and-place task (Section~\ref{sec:validation}). Finally, Section~\ref{sec:conclusion} summarizes the paper and gives an outlook.

\section{Related Work}
\label{sec:related_work}
The usage of VLAs for robotics is a relatively new field with high traction. VLAs are a core innovation for humanoid robots, allowing non-deterministic, adaptive behavior in diverse scenarios. However, VLAs are also finding their way into traditional serial kinematics~\cite{Kawaharazuka2025}.
Early iterations of language-conditioned policies, such as RT-1~\cite{Brohan2023}, relied on routing independently tokenized embeddings from efficient convolutional or vision transformer networks through standard transformer blocks to output discretized control commands. While effective within structured domains, these architectures exhibited limited open-vocabulary generalization.

\subsection{Vision-Language-Action Models and Their Usage in Industrial Environments}
The state of the art shifted significantly with the introduction of models like RT-2~\cite{Zitkovich2023} and the Open X-Embodiment initiative~\cite{openxembodimentcollaboration2025}. These architectures directly co-train VLM backbones (such as PaLI-X~\cite{chen2023} or \mbox{PaLM-E}~\cite{driess2023}) on web-scale VLM tasks and physical robotic trajectories.
By phrasing robotic action generation as a token-prediction problem, these models naturally inherit semantic reasoning capabilities. For instance, given a scene with novel objects and a natural-language instruction, the VLA leverages its knowledge to identify the object before mapping the spatial coordinates to end-effector actions via its embedded policy~\cite{Zitkovich2023,openxembodimentcollaboration2025}.
Building on this paradigm, OpenVLA~\cite{Kim2024} introduced an open-source 7B parameter generalist policy trained natively on the Open X-Embodiment dataset. By fusing open-weight vision-language backbones with robust action tokenization, it democratized access to state-of-the-art cross-embodiment control.
Cross-embodiment control refers to single policies enabling the control of diverse robot kinematics (embodiments), overcoming the limitations of traditional control engineering that has to be designed specifically for a kinematic structure~\cite{Yang2024}.
Octo~\cite{octomodelteam2024}, also trained on X-Embodiment, couples a block-wise masked attention pattern, that allows flexibly adding or removing observations during finetuning, with a lightweight diffusion-based action head. By this, Octo allows lenient adaptation towards cross-embodiments and different robot/sensor setups. With 27M (Octo-Small) or 93M (Octo-Base) parameters, finetuning and inference are computable on consumer-grade hardware. Notably, Ghosh et al.~\cite{octomodelteam2024} report comparable results to much larger models like RT-2-X (55B)~\cite{openxembodimentcollaboration2025}.

While VLAs demonstrate remarkable scaling in laboratory settings, deploying them within high-throughput production environments introduces severe constraints.
First, standard automation systems demand fast control frequencies. Particularly in VLAs with many billion parameters, inference may induce latencies above an acceptable threshold.
On the one hand, current research overcomes this limitation through aggressive quantization, tensor parallelism, and action-chunking mechanisms where the model outputs a sequence of future actions in a single inference step~\cite{Song2026,Black2025}.
On the other hand, VLAs with fewer parameters may be used that lead to less latency through faster inference.
%To satisfy this demand, recent architectures compress parameter boundaries to achieve higher control frequencies natively on edge devices.
For instance, MiniVLA~\cite{Belkhale2024}, a smaller variant of OpenVLA, scales down the large footprint to just 1B parameters by pairing a compact Qwen language backbone with vector-quantized action chunking.
Complementing this trend, Hugging~Face introduced SmolVLA~\cite{Shukor2025}, a highly lightweight 450M-parameter foundation model optimized for deployment on consumer-grade hardware and CPUs.
%By strategically skipping the upper layers of its SmolVLM-2 base and utilizing pixel-shuffling to compress visual inputs to just 64 tokens per frame, SmolVLA significantly cuts computational overhead while retaining competitive cross-embodiment manipulation performance.
Similarly, small models like PokeVLA~\cite{Zheng2026} demonstrate that optimizing world-knowledge guidance and visual tokens can effectively offset a reduced language-model capacity, allowing high-frequency real-time execution without sacrificing generalization.

Second, most VLAs are highly demanding on hardware, particularly GPUs that are not necessarily available on the edge. Outsourcing compute to proprietary cloud services raises compliance and privacy issues.
On the one hand, industrial adopters may decide to use smaller VLAs that are less demanding on compute and can be computed with edge devices. However, smaller models offer less semantic generalization capabilities and require more specialized deployment~\cite{Shukor2025}. 
On the other hand, instead of fully training the model, only the action head may be fine-tuned while keeping the backbone frozen. This reduces the demand on compute for model training significantly~\cite{Kawaharazuka2025}. In addition, Low-Rank Adaptation (LoRA)~\cite{Yang2025} can be employed to insert parameter-efficient adapters into the attention and multi-layer perceptron layers of the pre-trained vision-language backbone~\cite{Kim2024,DonghoonKim2026}. By training only a fraction of the total parameters, this method lowers VRAM overhead significantly to allow multi-billion-parameter VLA optimization on consumer-grade hardware while preserving the backbone's crucial zero-shot semantic capabilities.

Third, smaller VLAs are mostly tested and benchmarked in labs with simple and small hardware, like SO-101 or OpenMANIPULATOR.
%~\cite{so101}.
%or OpenMANIPULATOR~\cite{Wie2026}. 
Particularly, SO-101 has been shown to be a suitable test bench for VLAs and robot learning~\cite{so101,Almuzairee2026}. However, this obscures the usage of the benchmarked VLAs with real industrial-grade hardware as control loops, safety requirements, payloads, and workspace dimensions vary significantly.

In conclusion, VLAs are a promising method for generalization and democratization of robot learning that are at the border of widespread adoption. However, the high latency and compliance concerns related to cloud-based training and inference should move adopters to models with fewer parameters that are computable on-premise and on the edge. These smaller VLAs have not yet been comprehensively benchmarked in industrial settings under the constraints of a real production environment. We aim to close this gap by allowing SmolVLA to be deployed on industrial-grade lightweight robots and by indicating learnings from moving away from purely lab- or consumer-grade hardware.

\subsection{SmolVLA}
\label{sec:smolvla}
SmolVLA by Shukor et al.~\cite{Shukor2025} is a lightweight VLA model (450M), designed explicitly to democratize VLA training and making it more accessible. The core innovation of SmolVLA lies in its computational efficiency, enabling on-premise execution and down-stream fine-tuning on consumer-grade hardware.
The structural backbone of SmolVLA is the pre-trained VLM SmolVLM-2~\cite{Marafioti2025}, which is natively optimized for multi-image and sequential video inputs. By incorporating diverse efficiency mechanisms -- most notably token-shuffling -- its vision encoder reduces redundant spatial representations. For multi-modal fusion, sequential video frames, natural language task descriptions, and robot sensorimotor states are tokenized into a unified discrete embedding space. Afterwards, they are concatenated sequentially into a single multi-modal context window.
To enforce real-time inference, the tokenization of input videos in SmolVLA is strictly bounded to 64 visual tokens per frame. Furthermore, the Action Expert is decoupled from the final layers of the network. By implementing a layer-skipping protocol, the architecture establishes an adjustable tuning parameter that explicitly balances between task performance (fewer skips) and computational efficiency (more skips). Layer-skipping attends the Action Expert only on the first $N$ layers of the VLM backbone. Shukor et al.~\cite{Shukor2025} observe that the later layers provide significantly less features and may be optimized similar to Pareto frontiers.

Departing from conventional VLA architectures that rely uniformly on full self-attention or homogeneous cross-attention mechanisms across all layers, SmolVLA introduces an interleaved attention topology. In this configuration, individual Transformer layers alternate between self- and cross-attention blocks, minimizing computational complexity while retaining global contextual awareness. Finally, the downstream action vector generation is formulated via a Flow Matching Transformer architecture, allowing the policy to model complex, multi-modal action distributions through continuous-time normalizing flows rather than discrete tokenization or mean-squared-error regression.
SmolVLA is pre-trained on community-driven trajectories aggregated under the \emph{lerobot} label on Hugging~Face. At submission, this repository comprised 1,270 heterogeneous datasets spanning a wide variety of robot embodiments, manipulation tasks, and environmental conditions~\cite{Shukor2025}. To ingest these diverse data, the training pipeline homogenizes proprioceptive dimensions and action spaces into a standardized coordinate framework. This extensive, crowd-sourced multi-task dataset ensures that despite its compact 450M parameters, SmolVLA retains strong out-of-distribution generalization capabilities and robust zero-shot transfer performance across novel environments.

\section{ROS2SmolVLA: Adapting SmolVLA for Industrial-Grade Lightweight Robots}
\label{sec:ROS~2smovla}
In an effort to enable industrial-grade lightweight robots to use SmolVLA, we propose ROS2SmolVLA, a ROS~2-based software interface, trained and validated with data of a Universal Robots UR10e. While the UR10e functions as a validation experiment, the methods are generalizable allowing cross-embodiment control partially without re-training the model.
ROS2SmolVLA draws from Hugging~Face's LeRobot framework~\cite{Cadene2026,Cadene2024} and the Cartesian Motion Controller by Scherzinger, Roennau, and Dillmann~\cite{Scherzinger2017}. In addition, the Robotiq Hand-E driver by the Center of Excellence in Artificial Intelligence (University of Krakow)~\cite{robotiq-hande-driver} is used.
ROS2SmolVLA uses following self-developed and self-recorded resources:
\begin{itemize}
    \item \href{https://github.com/una-auxme/ros2smolvla_docker}{\texttt{ros2smolvla\_docker}}: Docker container for the integrated hardware-software setup.
    \item \href{https://github.com/una-auxme/ros2smolvla_ur10e_sim}{\texttt{ros2smolvla\_ur10e\_sim}}/\href{https://github.com/una-auxme/ros2smolvla_ur10e_real}{\texttt{\_real}}: Execution on the robot featuring a virtual (\emph{sim}) and a real twin (\emph{real}).
    \item \href{https://github.com/una-auxme/ros2smolvla_interface_camera}{\texttt{ros2smolvla\_interface\_camera}}: ROS~2 camera integration for LeRobot.
    \item \href{https://github.com/una-auxme/ros2smolvla\_interface\_lerobot}{\texttt{ros2smolvla\_interface\_lerobot}}: ROS~2 interface for LeRobot/SmolVLA.
    \item Hugging~Face data and model repository including \href{https://huggingface.co/datasets/una-auxme/ROS2SmolVLA_ur10e_no_joints_crop_pick_place}{training} and \href{https://huggingface.co/datasets/una-auxme/eval_ros2smolvla}{validation} data, and the \href{https://huggingface.co/una-auxme/ROS2SmolVLA_ur10e_no_joints_crop_pick_place}{model}~\cite{auxme:ros2smolvla_huggingface}.
\end{itemize}
All resources are accessible through the docker repository~\cite{auxme:ur10e-vla-docker} and listed on a central Github webpage which we recommend as an entry point: \href{https://una-auxme.github.io/en/projects/ros2smolvla/}{https://una-auxme.github.io/en/projects/ros2smolvla/}

\subsection{System Architecture and ROS~2 Interface}
\label{ssec:ROS2SmolVLA_architecture}
\begin{figure}[htb]
    \centering
    \includegraphics[width=.87\linewidth]{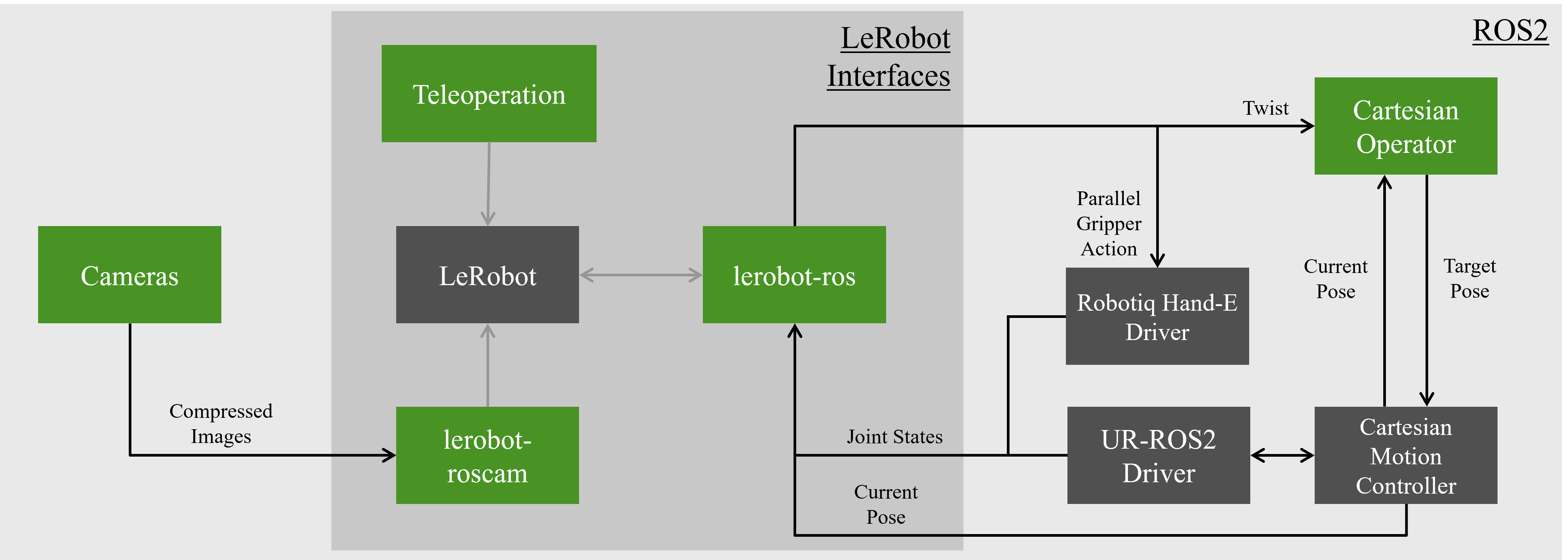}
    \caption{Architecture of ROS2SmolVLA. Self-developed software (green) and external dependencies (gray) are marked as boxes. Arrows indicate communication (dark: ROS~2 topics, light: LeRobot interface).}
    \label{fig:architecture}
\end{figure}

The architecture of ROS2SmolVLA utilizes a decentralized edge-to-workstation paradigm designed to decouple real-time hardware execution from resource-intensive multi-modal inference. The communication layer relies on \texttt{lerobot-ros}, acting as a lightweight broker between the LeRobot and ROS~2 ecosystems. This abstraction enables the core VLA policies to seamlessly operate on raw joint trajectories or Cartesian velocity spaces without managing low-level industrial bus cycles directly. Figure~\ref{fig:architecture} sketches the interactions between the system's software components.
At the hardware execution layer, the robot arm's actuation modes are abstracted through standardized \texttt{ros2\_control} controllers. Depending on operational requirements, the system exposes four ways of interacting with the robot: direct joint position control (\texttt{position\_controllers/JointGroupPositionController}), joint trajectory interpolation (\texttt{joint\_trajectory\_controller}), real-time operational-space velocity tracking utilizing \texttt{MoveIt Servo}, and publishing a \texttt{TwistStamped} message for ingestion by the Cartesian Motion Controller through our Cartesian Operator. For the Robotiq Hand-E end-effector, gripper commands are processed concurrently via either a trajectory-based controller or a feedback-oriented \texttt{(Parallel)GripperActionController}. State tracking and feedback verification are broadcast across the local ecosystem using a centralized \texttt{joint\_state\_broadcaster} and the Cartesian Controllers' \texttt{current\_pose} topic for the current Cartesian pose of the end-effector. By utilizing Cartesian coordinates instead of direct joint control, our approach can easily be ported to different embodiments. 

\subsection{Multi-Modal Data Acquisition}
\label{ssec:ROS2SmolVLA_data}
To train and guide SmolVLA, a data collection and sensor aggregation layer is established across the physical and a virtual (Gazebo) platform. Visual state representation is driven by \texttt{lerobot-roscam}, a dedicated self-developed ROS~2 data-acquisition package designed to ingest asynchronous camera inputs. \texttt{lerobot-roscam} standardizes the ingestion pipelines for the workspace's two static cameras and the end-effector-mounted camera, by implementing the LeRobot camera interface and converting camera signals from ROS~2 messages thereof. This matches the multi-modal requirements of the base VLM model. Optionally, \texttt{lerobot-roscam} can also augment the incoming image by converting the color space, rotating the image, and cropping to custom resolutions.
During operation, the LeRobot framework utilizes \texttt{lerobot-ros} for connecting to the robot, sending commands, and collecting the proprioceptive state. LeRobot then merges the high-dimensional visual tensors obtained from \texttt{lerobot-roscam} with low-dimensional robot telemetry obtained from \texttt{lerobot-ros}. Textual instructions or semantic language prompts are concurrently mapped to these multi-modal states within continuous action episodes. The target trajectories (joint velocities or Cartesian paths) and the normalized jaw openings are indexed into serialized data records matching LeRobot's standardized format. This allows the edge device to serialize datasets directly from the active industrial network.

\subsection{Containerized Inference and Simulation Environment}
\label{ssec:ROS2SmolVLA_container}
The final layer of the architecture focuses on isolating dependencies, scaling model execution, and offering digital twin capabilities. To achieve consistent reproducibility between development machines, edge devices and production workstations, the software environment is packaged via \texttt{ur10e\_vla\_docker}. This containerized environment isolates the CUDA runtime, PyTorch extensions, and ROS~2 packages via Docker~\cite{miell2019}. On the inference workstation, this containerized wrapper manages and converts the incoming state data, feeds them to the SmolVLA network, and computes the corresponding joint or velocity action targets at low latencies.
Complementing the physical deployment stack, \texttt{ur10e\_vla\_sim} acts as an architectural verification and simulation layer. The simulation environment replicates the physics, camera configurations, and workspace layout of the physical robot cell inside Gazebo. As the underlying ROS~2 control API and topic schemas within \texttt{ur10e\_vla\_sim} (virtual twin) are entirely identical to the physical setup handled by \texttt{lerobot-ros} and \texttt{ur10e\_vla\_real} (the real twin), the containerized SmolVLA policies can be deployed interchangeably. This structural parity enables pre-validation and fine-tuning of the VLA policy within a simulated context prior to target execution on the physical UR10e system. Further, hybrid training data may be generated and combined from the simulated and real environment. While we integrated the simulation capabilities into ROS2SmolVLA, we utilized only real data in later validation. Therefore, the simulator was not fully validated.

\section{Validation}
\label{sec:validation}
Validation aims to verify the viability of SmolVLA in combination with industrial-grade hardware. We feature a simple pick-and-place scenario in which the robot is tasked to insert cubes into boxes.
This scenario, albeit academic, is generalizable toward industrial environments, as pick-and-place tasks with small items are common. Within nine scenarios, we test diverse influential factors, including colors, positions, configurations, geometries, and object count.

\subsection{Reference Hardware, Data, and Model Training}
\label{ssec:validation_data}
Our hardware setup (Figure~\ref{fig:workbench_overview}) features two Microsoft Azure Kinect cameras, a webcam mounted to the end-effector, and a Robotiq Hand-E jaw gripper. The cameras' data streams are processed and concatenated on a Nvidia Jetson AGX Orin edge PC. The VLA is computed on a workstation with consumer-grade hardware (Nvidia RTX~4080, 16GB VRAM, 32GB RAM). The distributed system is connected via an unmanaged 10Gbit Ethernet switch.
\iffalse
\begin{figure}[htb]
    \centering
    \includegraphics[width=0.65\linewidth]{figures/workbench_overview_cropped2.jpg}
    \caption{Hardware setup used for the validation and data generation of ROS2SmolVLA in a lab environment. The main robot workspace is on the left half. The scene's lighting is well-controlled through blinders, a diffuse top light and two flicker-less side lights. As the hardware is used in multiple projects, additional tools and markings are present. These are used as interferences for the target process.}
    \label{fig:hardware}
\end{figure}
\fi
Our datasets contain three camera views, the joint states, and the Cartesian end effector position to maximize flexibility on input features for the model. We chose Cartesian delta values as actions, which are also included in the data.
Training data was generated semi-structured via teleoperation. The robot and its end effector position were controlled with a standard Gamepad. The recorded episodes contain the full action sequences from the robot start configuration to cube insertion into the box. Between each repetition, the robot was manually moved back to the starting position. This resulted in slight variations of the robot starting configuration. To learn the spatial relationships of the workspace, a 5x5 cube position grid was established. For each cube position five repetitions with varied target box positions were recorded. In addition, episodes that contain multiple cubes with different colors were introduced. To enable recovery after erroneous grasps or dropped cubes, episodes showing error-state recovery patterns were also included in the data set.
The dataset with a total of 349 episodes is available via Hugging~Face~\cite{auxme:ros2smolvla_huggingface}.
%In total, 349 episodes are represented.
%In the dataset, we varied cube color, cube count, robot configuration, and positions of cubes and the box.

%\subsection{Model Training}
%\label{ssec:validation_model_training}
The model was fine-tuned following the instructions provided in the LeRobot documentation using the pretrained base model by Shukor et al.~\cite{Shukor2025}. Particularly, we set \texttt{train\_expert\_only=False} and \texttt{compile\_model=True} which resulted in better learning performance. Otherwise, default parameters were used. While training runs were initially stopped after 50,000 steps, we noticed that training for 200,000 steps improved the success rates significantly, while being more demanding on compute. Therefore, we moved the training of the 200,000-steps-model to a GPU server with a NVIDIA L40S. Training on the server took about 25~hours. Note that the model can also be trained on the workstation, but training will take significantly longer.
Our trained policy uses the Cartesian end effector pose and the three visual features as inputs. To make efficient use of the restricted visual token count per image, we cropped the top view to the lower half which contains all of the relevant work space (cf. Figure~\ref{fig:workbench_overview}). The outputs of the model are \texttt{TwistStamped} commands which then get applied to the robot's end effector position.

\subsection{Test Plan and Results}
\label{ssec:validation_plan}
\begin{figure}[ht]
    \centering
    \resizebox{\linewidth}{!}{%
        \includegraphics[width=.594\linewidth]{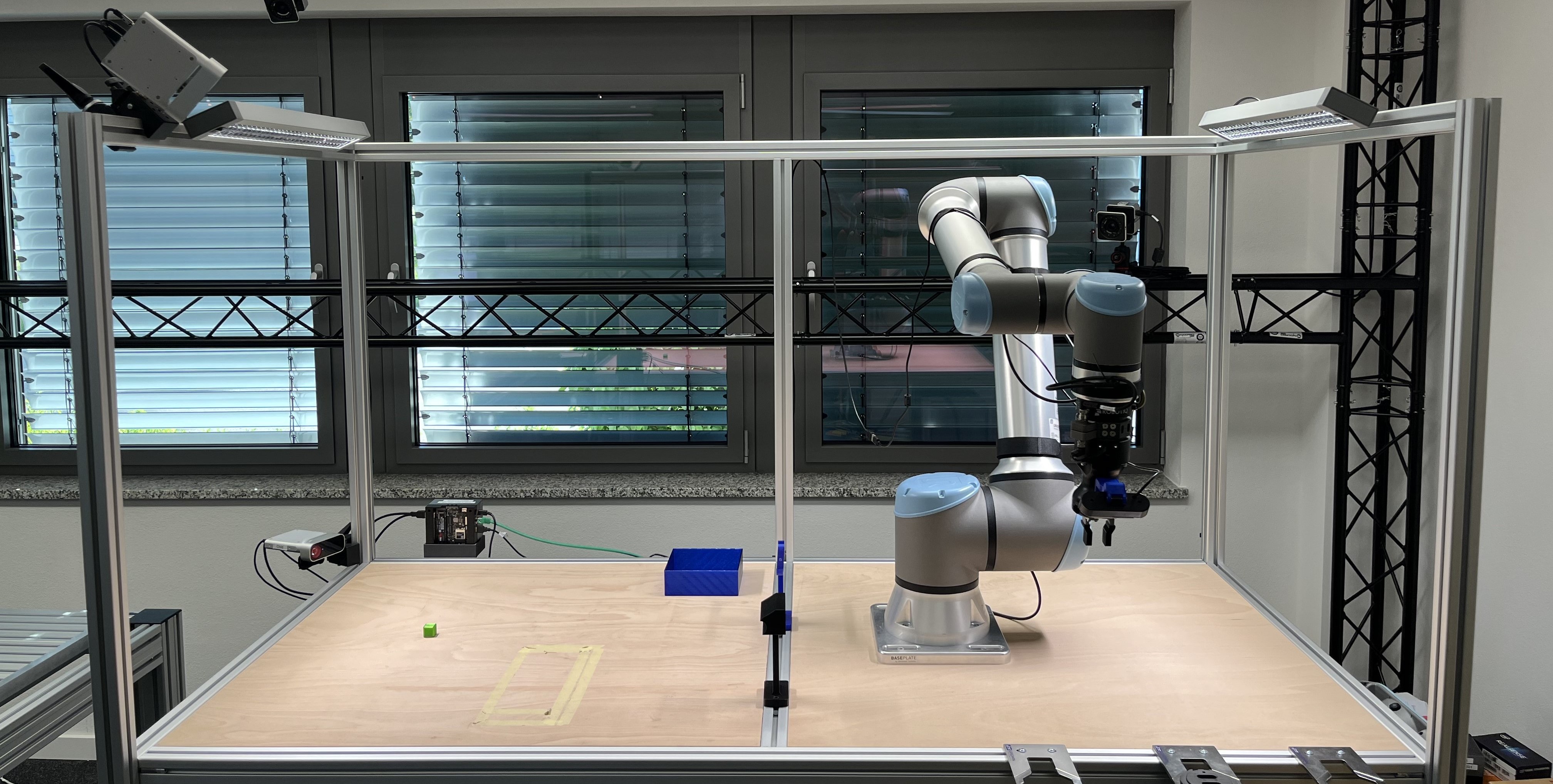}\hspace*{1mm}%
        \includegraphics[width=.4\linewidth]{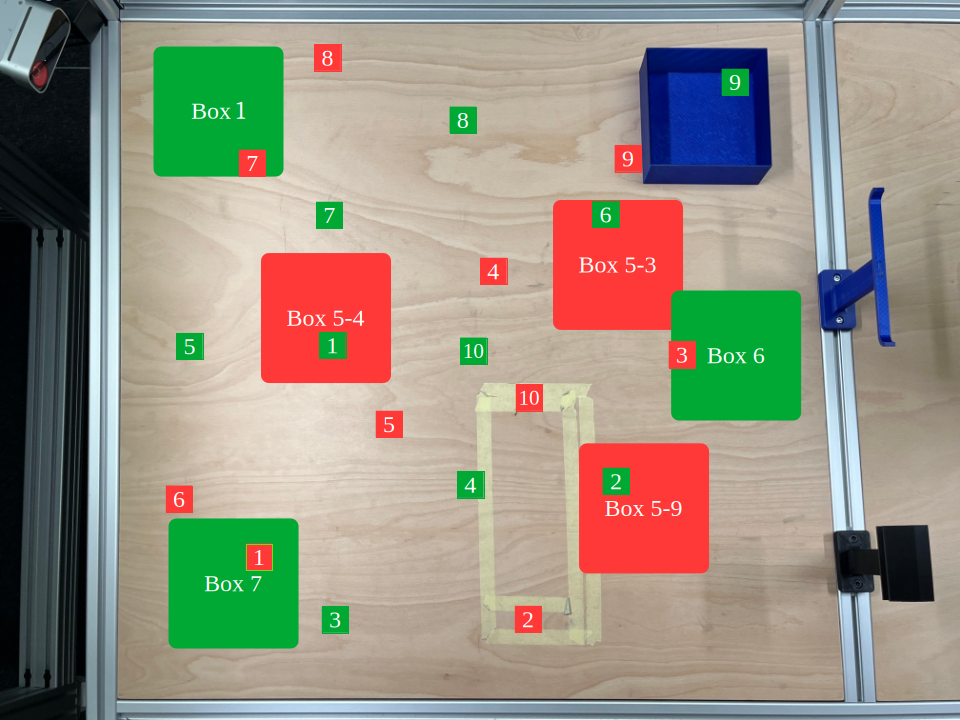}
    }
    \caption{Left: Hardware setup used in this paper. The main robot workspace is on the left half. The scene's lighting is well-controlled through blinders, diffuse top lights and flicker-less side lights. As the hardware is used in multiple projects, additional tools and markings are present. These are used as interferences for the target process. Right: ID (green) and OOD (red) CP, and BP in the robot's main workspace.}
    \label{fig:workbench_overview}
\end{figure}

The test plan consisted of nine structured scenarios. We use the shortened scripts: cube position (CP), cube color (CC), box position (BP), box color (BC), box shape (BS), robot configuration (RC), in-distribution (ID), out-of-distribution (OOD), and test scenario (T). The following test cases were defined before testing:
\begin{itemize}
    \item[\textbf{T1}] Variation of ID CP (10~variants) and ID BP (3~variants).
    \item[\textbf{T2}] Variation of OOD CP (10~variants) and OOD BP (3~variants).
    \item[\textbf{T3}] Variation of ID CP (10~variants) and RC (6~variants).
    \item[\textbf{T4}] Variation of ID CC (3~variants) and ID CP (10~variants).
    \item[\textbf{T5}] Single OOD CC and variation of ID CP (10~variants).
    \item[\textbf{T6}] Two different-colored cubes at ID CP (10~variants) with varied order (2~variants); single cube picked without indicating exact color in prompt.
    \item[\textbf{T7}] Two different-colored cubes at in-distribution CP (10~variants) with varied order (2~variants); single colored cube picked according to color indicated in prompt.
    \item[\textbf{T8}] Variation of OOD BC (2~variants) and ID BP (6~variants).
    \item[\textbf{T9}] Variation of OOD BS (2~variants) and ID BP (6~variants).
    % \item[\textbf{E}] CP and BP at workspace limits, and other edge cases.
\end{itemize}
Due to the way data was generated, there is no clear distinction in ID and OOD for RC.
We divided the workplace into nine uniform zones and placed markers at each center on which BP were varied. As randomized BP were also used in few training data, there is no clear distinction in ID and OOD for BP. However, we chose similar ID BP to the training data.
Figure~\ref{fig:workbench_overview} indicates the ID and OOD CP/BP.
All tests used the prompt \emph{``Pick up the [CC] cube and put it in the [BC] box.''}. For T6, the [CC] modifier was omitted.
%\begin{figure}[htb]
%    \centering
%    \includegraphics[width=0.55\linewidth]{figures/Box_Dice_positions.png}
%    \caption{ID (green) and OOD (red) cube and box positons on the robots main workspace.}
%    \label{fig:box_dice_pos}
%\end{figure}

%\subsection{Results}
%\label{ssec:validation_results}
For reporting, we split the pick-and-place action into pick and place sub-tasks, i.~e. if the cube was picked and kept gripped among the whole trajectory, and if the cube was placed into the box, respectively. If the robot recovered automatically, we counted the attempt as successful, e.~g., if the robot dropped the cube but managed to pick it up again. Place scores are reported from total, hence, failed picks also result in failed place sub-actions. The evaluation dataset is available on Hugging~Face~\cite{auxme:ros2smolvla_huggingface}.
Table~\ref{tab:t1_t5} reports the variations of BP, CP, RC, and CC (T1-T5).
\begin{table}[htb]
    \centering
    \caption{Test cases T1 to T5 with evaluation of pick and place actions. Left: Variations of BP and CP (OOD BP between two zones indicated by two IDs). Middle: Variations of RC, tested with ID CP and BP Box 7. Right: ID and OOD variation of CC, tested with ID CP and BP Box 7.}
    \label{tab:t1_t5}
    \begin{tabular}[t]{m{3mm} c c c}
        & \textbf{BP} & \textbf{Pick} & \textbf{Place} \\
        \cline{2-4}
        \multirow{3}{*}{\textbf{T1}} & 1 & 90\% & 90\% \\
        & 7 & 100\% & 100\% \\
        & 6 & 100\% & 40\% \\
        \cline{2-4}
        \multirow{3}{*}{\textbf{T2}} & 5-4 & 70\% & 60\% \\
        & 5-9 & 90\% & 70\% \\
        & 5-3 & 40\% & 20\% \\
    \end{tabular}%
    \hspace*{5mm}
    \begin{tabular}[t]{m{3mm} c c c}
        & \textbf{RC} & \textbf{Pick} & \textbf{Place} \\
        \cline{2-4}
        \multirow{6}{*}{\textbf{T3}} & 1 & 70\% & 70\% \\
        & 2 & 80\% & 80\% \\
        & 3 & 90\% & 80\% \\
        & 4 & 80\% & 80\% \\
        & 5 & 50\% & 50\% \\
        & 6 & 90\% & 90\% \\
    \end{tabular}%
    \hspace*{5mm}
    \begin{tabular}[t]{m{3mm} c c c}
        & \textbf{CC} & \textbf{Pick} & \textbf{Place} \\
        \cline{2-4}
        \multirow{3}{*}{\textbf{T4}} & Yellow & 50\% & 50\% \\
        & Blue & 60\% & 60\% \\
        & Gray & 80\% & 80\% \\
        \cline{2-4}
        \textbf{T5} & Orange & 60\% & 60\% \\
    \end{tabular}
\end{table}

In \textbf{T6} picks were repeated until a cube was gripped to indicate which cube was chosen and whether CC influenced its manipulation. The model was presented a green and gray cube. Under this preliminaries, the model chose the green cube 75\% of all repetitions. Place sub-actions always succeeded.
\iffalse
\begin{table}[!h]
    \centering
    \begin{tabular}{m{3mm} c c c c}
        & & \textbf{Pick} & \textbf{Green Pick} & \textbf{Place} \\
        \hline
        \multirow{2}{*}{\rotatebox[origin=c]{90}{\textbf{T6}}} & 2 Color 1& 100\% & 80\% & 100\%\\
        & 2 Color 2& 100\% & 70\% & 100\% \\
    \end{tabular}
    \caption{CC comparison without color in prompt}
    \label{tab:t6}
\end{table}
\fi
For T7, we essentially repeated T6 with additionally indicating a desired CC in the prompt. We report the compliance with the color and the place rate in Table~\ref{tab:t7}.
\begin{table}[htb]
    \centering
    \caption{CC comparison with color indicated in prompt.}
    \label{tab:t7}
    \begin{tabular}{m{3mm} c c   c c}
        & \textbf{Cubes present} & \textbf{Prompt} & \textbf{Correct color picked} & \textbf{Place} \\
        \cline{2-5}
        \multirow{4}{*}{\textbf{T7}} & 2 & green & 70\% & 100\%\\
        & 2 & gray & 20\% & 90\% \\
        & 2, CPs inverted & green & 70\% & 100\% \\
        & 2, CPs inverted & gray & 30\% & 90\% \\
    \end{tabular}
\end{table}

Table~\ref{tab:t8_t9} reports on the variations in BC and BS. In T8, we presented the model with a black and orange square box that were not part of training data. In T9, we presented the model with a blue round box and a cardboard box. Latter varies not only in shape but also introduces a completely new texture.
\begin{table}[htb]
    \centering
    \caption{OOD box colors and shapes. Left: Variation of OOD BC. Right: Variation of OOD BS.}
    \label{tab:t8_t9}
    \begin{tabular}[t]{m{3mm} c c c c}
        & \textbf{BC} & \textbf{Pick} & \textbf{Found box} & \textbf{Place} \\
        \cline{2-5}
        \multirow{2}{*}{\textbf{T8}} & Black & 100\% & 17\% & 17\% \\
        & Orange & 100\% & 100\% & 17\% \\
    \end{tabular}%
    \hspace*{5mm}
    \begin{tabular}[t]{m{3mm} c c c c}
        & \textbf{BS/BC} & \textbf{Pick} & \textbf{Found box} & \textbf{Place} \\
        \cline{2-5}
        \multirow{2}{*}{\textbf{T9}} & Round Blue & 100\% & 100\% & 100\% \\
        & Cardboard & 83\% & 50\% & 17\% \\
    \end{tabular}
\end{table}
\iffalse
\begin{table}[htb]
    \centering
    \caption{OOD box colors and shapes}
    \label{tab:t8_t9}
    \begin{subtable}[t]{0.49\textwidth}
        \centering
        \caption{Variation of OOD BC.}
        \label{tab:t8}
        \begin{tabular}{m{2mm} c c c c}
            & \textbf{BC} & \textbf{Pick} & \textbf{Found box} & \textbf{Place} \\
            \hline
            \multirow{2}{*}{\textbf{T8}} & Black& 100\% & 17\% & 17\%\\
            & Orange& 100\% & 100\% & 17\% \\
        \end{tabular}        
    \end{subtable}
    \begin{subtable}[t]{0.49\textwidth}
        \centering
        \caption{Variation of OOD BS.}
        \label{tab:t9}
        \begin{tabular}{m{2mm} c c c c}
            & \textbf{BS/BC} & \textbf{Pick} & \textbf{Found box} & \textbf{Place} \\
            \hline
            \multirow{2}{*}{\textbf{T9}} & Round Blue & 100\% & 100\% & 100\% \\
            & Cardboard & 83\% & 50\% & 17\% \\
        \end{tabular}        
    \end{subtable}
\end{table}
\fi

Lastly, we report the \textbf{overall scores} as success of [pick, place, place after successful pick], respectively. In the ID cases (T1, T3, T4), the success rates were [78.33\%, 72.50\%, 92.47\%]. In the OOD cases (T2, T5, T8, T9), the success rates were [76.56\%, 46.88\%, 61.22\%]. Among all test cases the success rates were [77.72\%, 63.59\%, 81.69\%].

\iffalse
Lastly, Table~\ref{tab:total_scores} lists the accumulated success rates for picks and places, also including PASP. For pick rates, T6 and T7 were omitted.
\begin{table}[htb]
    \centering
    \caption{Total success rates among all test scenarios.}
    \label{tab:total_scores}
    \begin{tabular}{c c c c}
        \textbf{Data type} & \textbf{Pick} & \textbf{Place} & \textbf{PASP}\\
         \hline
        ID (T1, T3, T4)         & 78.33\% & 72.50\% & 92.47\%\\
        OOD (T2, T5, T8, T9) & 76.56\% & 46.88\% & 61.22\%\\
        \hline
        \textbf{Overall}                              & 77.72\% & 63.59\% & 81.69\%
    \end{tabular}
\end{table}
\fi

%The pick data in table \ref{tab:id_ood_pick} represents all applicable recorded picks in the respective distribution category.

\iffalse
%%%%%%%%%% Overall Grasping & Placing %%%%%%%%%%
\begin{table}[!h]
    \centering
    \begin{tabular}{r|c c}
        %\multicolumn{2}{r|}{\textbf{Test Cases}} & \textbf{Pick} & \textbf{Place} \\
        & \textbf{Green only} & \textbf{Overall} \\
        \hline
        ID & 83\% & 77\%\\
        OOD & 66\% & --\%\\
        
    \end{tabular}
    \caption{Pick performance ID vs. OOD}
    \label{tab:id_ood_pick}
\end{table}
\fi

\subsection{Discussion}
\label{ssec:validation_discussion}
As indicated by T1 and T2, while the system performed reliably for most ID BP, specific spatial configurations led to lower success rates. For instance, in T1, while BP~1 and BP~7 yielded high success rates, the success of place sub-actions was significantly impaired at BP~6. This suggests a failure in spatial reasoning, even though in other experiments a similar box position, e.~g., in T8/T9, did not necessarily result in lower success rates. We consider this to be a cause of a lack of training data coverage for this region of the workspace. During grasping, we observed that the policy successfully learned to grip the cube along the faces and to turn the end-effector appropriately to avoid gripping the edges. As expected, OOD RCs (T2) showed a broader performance degradation, with BP~5-3 failing at both the pick and place sub-actions.
T3 highlighted a dependency on the initial RC. While most configurations maintained high success rates, RC~5 caused a drop. This is explained easily: RC~5 is located far outside the camera's starting envelope, limiting the wrist camera's effectiveness in early detection of the cube's location.

T4 and T5 revealed a distinct bias in our vision model by showing lower pick rates for yellow and orange compared to the gray cubes.
%This bias is further quantified in \textbf{T6}:
Further, when presented two cubes without a specific prompt (T6), the system defaulted to picking the green cube. More critically, T7 demonstrated a failure in prompt adherence for non-dominant colors as
the success rates significantly dropped when prompting for the gray cube.
%When explicitly prompted to pick the green cube, the system succeeded 70\% of the time, while success rate dropped to 20–30\% for the gray cube.
%. However, when prompted to pick the gray cube, accuracy dropped to 20–30\%.
The system effectively ignored the condition and defaults to its inherent bias for the green cube. Overall, the color distinction capabilities of the trained policy are negligible despite 33\% of the training episodes contain multiple colored cubes with explicit prompts to pick a certain color.
The OOD BC/BS tests T8 and T9 exposed limitations in how the system interprets receptacles. BC and BS drastically alter the success of place sub-actions. The detection of the black box mostly failed. Interestingly, the orange box was consistently detected, but the place sub-action still mostly failed, indicating that the policy uses blue color in the wrist camera to decide when to drop the cube. This was manually verified by holding a blue object in front of the camera. Conversely, the round blue box achieved a perfect success rate, showing that the place sub-action generalizes well to certain novel shapes, provided the visual contrast allows for accurate boundary detection as the input resolution of the VLM is relatively low. While the success rate for the cardboard box -- with significantly different shape and texture -- dropped considerably, this still indicates the model's ability to generalize to the concept of a box.

%When testing the cardboard box, with drastically different shape and texture, the policy's locating performance dropped, but still indicates the model's ability to generalize to the concept of a box rather than specific BS/BC.

Albeit these limitations, the accumulated scores indicate a general feasibility of SmolVLA with industrial-grade hardware. The OOD picks were on par with ID tests while the success of place sub-actions were significantly lowered, particularly due to the bias towards blue as trigger for dropping the cube.
Despite the results showing an expected gap between ID and OOD applications and the limitations of the model, the overall scores reflect a general ability of the model for generalization.
From the validation, we derived measures to improve data acquisition and model performance. With suitable improvements, we assume that SmolVLA will be a viable model architecture for small-scale applications in industrial environments.

%The aggregate data in table \ref{tab:id_ood_pick} confirms the expected gap between ID and OOD performance. The ID pick rate stands at 77\%, heavily buoyed by the baseline green cube (83\%). The drop to a 66\% OOD pick rate reflects the current boundaries of the system's generalization.

\subsection{Lessons Learned}
\label{ssec:validation}
While validation indicates that SmolVLA is a viable option for the deployment of VLAs on industrial-grade hardware, it is lacking robustness among holistic deployment scenarios. Thus, we indicate ``lessons learned'' that leverage the foundations made with ROS2SmolVLA and indicate useful extensions that will likely increase its robustness.

%\subsubsection{Model Inputs and Object Dimensions}
The model's low input image resolution impacts the efficiency in interpreting visual data. In initial tests, the model was unable to locate objects outside the wrist camera's view, likely due to the tokenization of the input images. After cropping the top view to the relevant workspace, locating the cube was significantly more successful. Thus, aggregating as much information as possible into a visual frame is crucial. This also applies to physical dimensions as very small objects may not be processed by the model at all.
In addition, the VLM has difficulties to find objects occluded by other task-relevant objects. In our case, the success rate dropped in scenarios where the cube was occluded by the box. This is likely due to the aggressive token optimization that plays a significant role in locating target objects.
In addition, the validation indicated a significant reliance of the model on colors and textures seen during training.
On the one hand, this indicates that training data must feature all relevant configurations of colors and textures that are expected in the workspace. Further, just adding more examples for these is not sufficient, as a dominant sample population may induce significant biases. In addition, the validation showed that the model uses blue as a primary trigger for place actions. Shape and edges are only used as secondary triggers. Generating more diverse training data may overcome this limitation and push edges towards the primary feature, leading to more generalization on the nature of the objects instead of the explicit instances.
On the other hand, this behavior could be exploited in application. Our model shows an inability to properly place objects in a black box due to an under-representation in the training data. Therefore, this could be an approach to specifically design for color-coded objects that the robot shall not interact with. This could be reinforced by specifically inserting negative conditioning cues into the training prompts.

%\subsubsection{Data Generation}
To improve generalization and recovery behavior, failure state recovery episodes can teach the policy how to behave in edge cases. We observed improved model behavior and greater robustness by adding such episodes.
However, during initial training these could bias the model towards unwanted behavior, e.~g., dropping the object before firmly gripping it. So, while failure state recovery episodes reinforce robustness in large data sets, they could lead to divergent behavior in sparse data.
Further, manual data generation is prone to introducing bias, e.~g., using similar trajectories to reach the target reproduces them in the model. In addition, teleoperation is labor intensive. There are approaches for synthetic data generation that might be applicable to the training of VLAs. We prepared the package \texttt{ur10e\_vla\_sim} for this purpose. However, it was not fully validated to produce comparable models to the ones trained with real data.

\section{Conclusion}
\label{sec:conclusion}
In this paper, we presented ROS2SmolVLA, a framework for integrating Hugging~Face's SmolVLA with industrial-grade lightweight robots via ROS~2. ROS2SmolVLA is composed of comprehensive components for containerization, wrapping SmolVLA, enabling synchronized vision inputs of multiple heterogeneous camera devices, and enabling actuation on the industrial-grade robot hardware and its digital twin via simulation in Gazebo. We trained our model for a pick-and-place task with 349 episodes and performed inference on compute-restricted hardware. The validation showed a general viability of the approach albeit success rates are currently too low for real-world deployment. This is mainly due to the still lack of training data. Code, data, and model parameters are available via a Github webpage (see Section~\ref{sec:ROS~2smovla}), Github repositories (i.~a.,~\cite{auxme:ur10e-vla-docker}), and Hugging~Face~\cite{auxme:ros2smolvla_huggingface}. From the validation and our experiences with SmolVLA, we derived lessons learned that shall guide adopters to overcome the limitations we currently encounter.
In future work depth information of the cameras could be leveraged. This was considered during the system design phase but was ultimately denied as depth information is currently not supported by the LeRobot dataset format. This could be a convenient way to add additional modalities that might be able to help with spatial reasoning. In addition, the influence of the recovery episodes during first training vs. later finetuning could be investigated.

\section*{Acknowledgements}
We like to thank Van Pham Nguyen for her support in data acquisition and the Bavarian Hightech-Agenda for funding this work within the AI Production Network Augsburg.
Gemini 3.5/3.7 Flash was used in drafting and editing.

\bibliographystyle{model3-num-names}
\bibliography{references}

\end{document}